\pdfoutput=1

\documentclass[11pt]{article}

\usepackage[final]{acl}

\usepackage{times}
\usepackage{latexsym}

\usepackage[T1]{fontenc}

\usepackage[utf8]{inputenc}

\usepackage{microtype}

\usepackage{inconsolata}

\usepackage{graphicx}
\usepackage{amsmath}
\usepackage{amsfonts}
\usepackage[linesnumbered,ruled]{algorithm2e}
\usepackage{tikz}
\usepackage{xcolor}
\usepackage{color}
\usepackage{pgfplots}
\usepackage{pgfplotstable} 
\pgfplotsset{compat=newest}
\usepackage{arydshln}
\usepackage{soul}
\usepackage{multirow}
\usepackage{subcaption}
\usepackage{caption}
\usepackage{enumitem}
\usepackage{arydshln}

\title{SQLong: Enhanced NL2SQL for Longer Contexts with LLMs}

\author{Dai Quoc Nguyen, Cong Duy Vu Hoang, Duy Vu, Gioacchino Tangari \\
{\bf Thanh Tien Vu, Don Dharmasiri, Yuan-Fang Li, Long Duong} \\
Oracle Corporation \\
{\tt{{\{dai.nguyen,vu.hoang,duy.vu,gioacchino.tangari\}@oracle.com}}} \\
{\tt{{\{thanh.v.vu,don.dharmasiri,yuanfang.li,long.duong\}@oracle.com}}}
}

\begin{document}
\maketitle

\begin{abstract}
Open-weight large language models (LLMs) have significantly advanced performance in the Natural Language to SQL (NL2SQL) task. 
However, their effectiveness diminishes when dealing with large database schemas, as the context length increases. 
To address this limitation, we present SQLong, a novel and efficient data augmentation framework designed to enhance LLM performance in long-context scenarios for the NL2SQL task. 
SQLong generates augmented datasets by extending existing database schemas with additional synthetic CREATE TABLE commands and corresponding data rows, sampled from diverse schemas in the training data. 
This approach effectively simulates long-context scenarios during finetuning and evaluation. 
Through experiments on the Spider and BIRD datasets, we demonstrate that LLMs finetuned with SQLong-augmented data significantly outperform those trained on standard datasets. 
These imply SQLong’s practical implementation and its impact on improving NL2SQL capabilities in real-world settings with complex database schemas.\footnote{Table Representation Learning Workshop at ACL 2025}
\end{abstract}

\section{Introduction}

The NL2SQL task focuses on translating natural language questions into SQL queries, enabling non-experts to interact with databases seamlessly \citep{deng-etal-2022-recent}. Recent advances leverage LLMs, finetuned on structured input prompts (\textit{e.g., task instructions, database schema, and natural language question}), to achieve state-of-the-art performance \citep{yang2024synthesizing,liu2024survey} on benchmarks such as Spider \citep{yu2018spider} and BIRD \citep{li2023can}.
Despite significant progress, a critical challenge persists: LLMs finetuned on existing benchmarks still struggle with large database schemas due to limited context handling. Current datasets primarily feature small schemas, failing to represent real-world complexities. Additionally, the absence of publicly available large-schema datasets further hinders progress. Addressing this, we propose SQLong, a data augmentation framework designed to enhance LLM performance in long-context NL2SQL tasks by extending schemas to meet predefined context thresholds.

\begin{figure}[t!]
\centering
\includegraphics[width=0.45\textwidth]{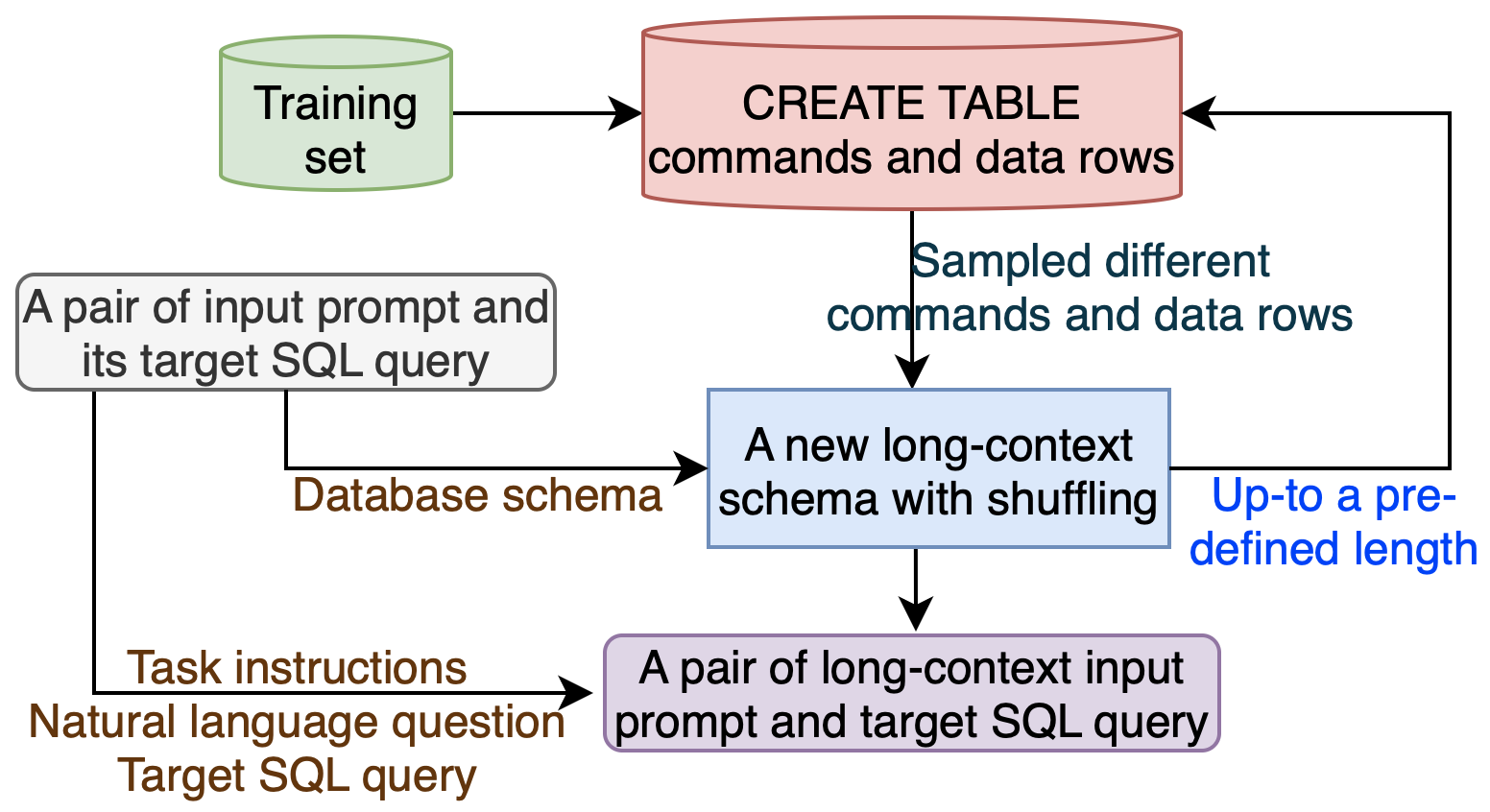}
\caption{Our proposed SQLong Pipeline.}
\label{fig:SQLong_pipeline}
\end{figure}

SQLong constructs augmented data by sampling $\mathsf{CREATE\ TABLE}$ commands and data rows from diverse schemas. These datasets enable LLMs to effectively manage large schemas and maintain robustness in long-context scenarios. Our experiments with \textit{CodeQwen1.5-7B-Chat} \citep{qwen} and \textit{Llama-3.1-8B-Instruct} \citep{dubey2024llama} show SQLong consistently outperforms baseline finetuning, achieving an average accuracy improvement of over 2.2\% on benchmarks like \texttt{Spider-dev}, \texttt{Spider-test}, and \texttt{BIRD-dev}.

Moreover, SQLong enables the creation of 45 long-context test sets, with context lengths up to 128k tokens. 
Models finetuned with SQLong exhibit significant performance gains, achieving an 11\% improvement over base models and a 6\% improvement over larger-scale models within the same family. These results highlight SQLong's effectiveness in real-world, large-schema scenarios.

In this paper, we focus on demonstrating that SQLong-augmented models outperform their unaugmented counterparts across varying context lengths. 
While direct comparisons to retrieval-augmented generation (RAG) schema linking are beyond this paper's scope, our findings suggest combining SQLong with RAG could unlock further gains. 
Our main contributions include:

$\bullet$ \textbf{Introducing long-context NL2SQL:} A challenging new task for evaluating LLM performance on large database schemas.

$\bullet$ \textbf{SQLong pipeline:} A novel, scalable data augmentation approach for generating long-context training and test datasets.

$\bullet$ \textbf{Empirical insights:} Comprehensive experiments validating SQLong's effectiveness in enhancing LLM robustness and accuracy in long-context scenarios.

$\bullet$ \textbf{Resource sharing:} Plans to release SQLong datasets and code to support further research.

\section{The Proposed SQLong Pipeline}
\label{sec:SQLong}
\begin{figure}[t!]
\centering
\includegraphics[width=0.42\textwidth]{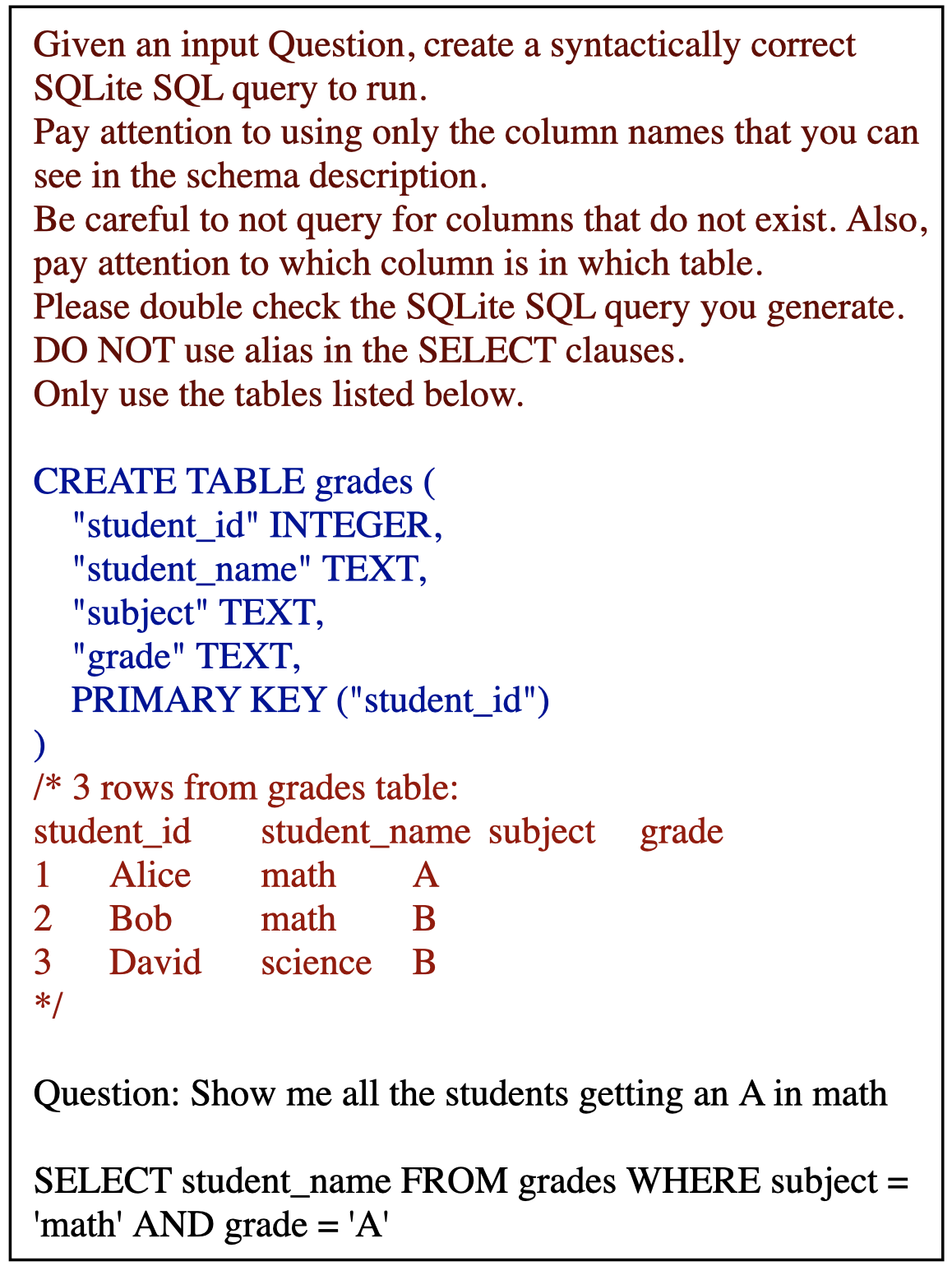}
\caption{Prompt template for the NL2SQL task.}
\label{fig:prompt_template}
\end{figure}
The NL2SQL task aims to translate a natural-language question about a database schema into a corresponding SQL query. 
Following the standardized prompt template \citep{rajkumar2022evaluating}, we represent the input prompt to LLMs in the format of \textit{(task instructions, database schema, natural language question)}.\footnote{In datasets with additional complexity, such as BIRD, the question may be supplemented with extra information, such as evidence. For simplicity, this additional information is omitted in Figure \ref{fig:prompt_template}.} 
As illustrated in Figure \ref{fig:prompt_template}, the database schema is represented by $\mathsf{CREATE\ TABLE}$ commands and three sample data rows for each corresponding table.

Using supervised finetuning (SFT) \citep{weifinetuned}, LLMs can be trained on pairs of input prompts and target SQL queries to optimize their performance on the NL2SQL task. 
Specifically, given a training set $\mathbf{T}$ comprising pairs of input prompts $\mathbf{x}$ and corresponding target SQL queries $\mathbf{s}$, the supervised finetuning process can be formulated as minimizing the log-likelihood loss \citep{weifinetuned}, as shown below:

$\mathbb{E}_{(\mathbf{x},\mathbf{s})\sim\mathbf{T}}\left[\sum_{i=1}^{|\mathbf{s}|}\log p_{\theta}\left(s_i | \mathbf{s}_{<i}, \mathbf{x}\right)\right]$

\noindent wherein $|\mathbf{s}|$ is the length of $\mathbf{s}$, $s_i$ is the $i$-th token, $\mathbf{s}_{<i}$ is the prefix of $\mathbf{s}$ up to the $i$-th position, and $\theta$ denotes the given LLM's parameters.

In this work, we introduce \textbf{SQLong}, a novel approach for constructing long-context finetuning and benchmark datasets, as illustrated in Figure~\ref{fig:SQLong_pipeline}. SQLong augments database schemas to enable large language models (LLMs) to effectively handle long-context scenarios in natural language to SQL (NL2SQL) tasks.

The SQLong pipeline has three main steps:


\textbf{1. Schema Collection.} We collect all $\mathsf{CREATE\ TABLE}$ commands and three sample data rows for each table from the training database schemas, compiling them into a comprehensive schema set.

\textbf{2. Schema Augmentation.} For each training pair, consisting of an input prompt (task instructions, database schema, natural language question) and its target SQL query, SQLong randomly samples items from the schema set. These sampled items contain table names distinct from those in the given database schema. The sampled items are combined with the original schema, and the resulting schema is randomly shuffled to produce a new, long-context database schema. This shuffling introduces variability in the positions of the original tables and columns.

\textbf{3. Long-Context Prompt Generation.} SQLong generates an augmented input prompt in the format of task instructions, the long-context database schema, and the natural language question, while keeping the target SQL query unchanged. It ensures that the combined length of the long-context input prompt and the target SQL query does not exceed a predefined context length (e.g., 32k tokens), maintaining compatibility with the model's tokenizer constraints.


By systematically extending and diversifying the context, SQLong enhances the robustness and effectiveness of LLMs in handling long-context NL2SQL tasks. We summarise the steps involved in SQLong in Algorithm \ref{alg:SQLong_algorithm} in Appendix \ref{subsec:SQLong_algorithm}.

\section{Evaluation}
We assess the effectiveness of our proposed SQLong model in enhancing NL2SQL performance in both short-context and long-context scenarios.

\subsection{Experimental Setup}
\label{subsec:exp_setup}

\paragraph*{\textbf{Datasets}} 
For the short-context evaluation, we utilize widely adopted benchmark datasets, including Spider \citep{yu2018spider}, Spider-realistic \citep{deng2020structure}, Spider-syn \citep{gan2021towards}, and BIRD \citep{li2023can}. \footnote{We use the latest BIRD-dev dataset, updated on June 27, 2024. The BIRD-test set is not publicly available.}
It is noted that Spider-Syn is manually created based on Spider training and development sets using synonym substitution in the original questions, while Spider-realistic is created based on Spider development set by manually removing the explicit mention of column names in the original questions. The BIRD-test set is not publicly available.

For the long-context evaluation, we extend each of the Spider-dev, Spider-test, Spider-realistic, Spider-syn, and BIRD-dev datasets by applying SQLong with a pre-defined context length. Specifically, we generate augmented long-context test sets for nine context lengths: 8k, 16k, 24k, 32k, 40k, 48k, 56k, 64k, and 128k. This process results in a total of 45 long-context test sets, constructed in accordance with the tokenizer of the base model.

Importantly, the long-context test sets are constructed with distinct database schema alignments. 
To build Spider-based long-context test sets, we use the database schemas from the BIRD training set, whereas for the BIRD-dev long-context test sets, we use the database schemas from the Spider training set. 
This ensures a robust evaluation across diverse schema configurations and context lengths.
The data statistics of the experimental datasets are presented in Figure \ref{fig:statistics_prompt_lengths_llama} and Tables \ref{tab:data_statistics} and \ref{tab:table_statistics}.

\begin{table}[!ht]
\centering
\includegraphics[width=0.235\textwidth]{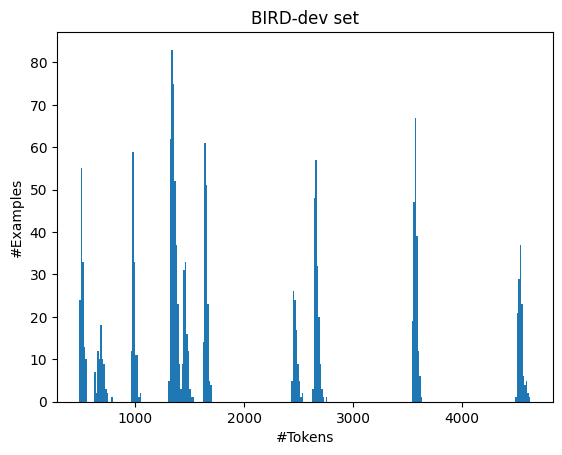}
\includegraphics[width=0.235\textwidth]{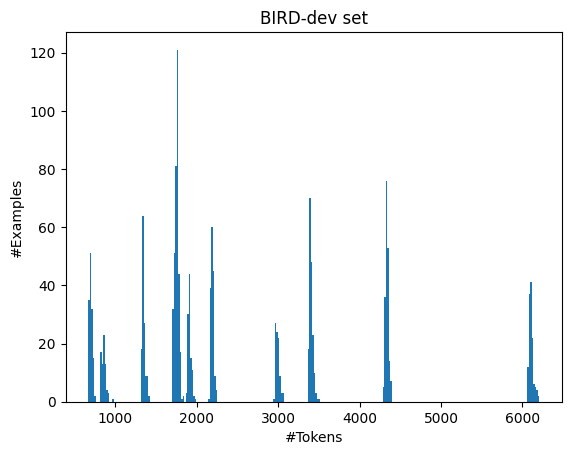}
\captionof{figure}{Statistics of input prompt lengths with respect to Llama-3.1-8B-Instruct's tokenizer (left) and CodeQwen1.5-7B-Chat's tokenizer (right) on the original BIRD-dev set. Similarly, the maximum input prompt lengths for the original Spider-related sets are approximately 2,000 tokens for Llama-3.1-8B-Instruct's tokenizer and 2,500 tokens for CodeQwen1.5-7B-Chat's tokenizer.}
\label{fig:statistics_prompt_lengths_llama}
\end{table}

\begin{table}[!ht]
\centering
\resizebox{7.5cm}{!}{
\def\arraystretch{1.05}
\begin{tabular}{l|lllll}
\hline
\bf Dataset & \bf \#DB & \bf \#tables & \bf \#training & \bf \#dev & \bf \#test\\
\hline
Spider & 200 & 5 $\pm$ 3 &  6,712 & 1,034 & 2,019\\
Spider-syn & 200 & 5 $\pm$ 3 & 6,712 & 1,034 & -- \\
Spider-realistic & 200 & 5 $\pm$ 3 & 6,712 & 508 & --\\
BIRD & 98 & 7 $\pm$ 3 & 9,428 & 1,534 & --\\
\hline
\end{tabular}
}
\caption{Statistics of the experimental datasets. \textbf{\#DB} denotes the number of databases. \textbf{\#tables} denotes the mean and standard deviation of numbers of tables in the databases.
}
\label{tab:data_statistics}
\end{table}

\begin{table}[!ht]
\centering
\resizebox{7.5cm}{!}{
\def\arraystretch{1.05}
\begin{tabular}{l|cc|cc}
\hline
\multirow{2}{*}{\bf Length} & \multicolumn{2}{|c}{\bf CodeQwen1.5-7B-Chat} & \multicolumn{2}{|c}{\bf Llama-3.1-8B-Instruct} \\
\cline{2-5}
 & \bf Spider-related & \bf BIRD-dev & \bf Spider-related & \bf BIRD-dev \\
\hline
8k & 37 $\pm$ 4 & 35 $\pm$ 8 & 48 $\pm$ 5 & 48 $\pm$ 8\\
16k & 72 $\pm$ 6 & 76 $\pm$ 8 & 94 $\pm$ 7 & 102 $\pm$ 9\\
24k & 107 $\pm$ 7 & 118 $\pm$ 8 & 141 $\pm$ 8 & 157 $\pm$ 9\\
32k & 142 $\pm$ 8 & 159 $\pm$ 9 & 186 $\pm$ 8 & 211 $\pm$ 9\\
40k & 177 $\pm$ 8 & 200 $\pm$ 9 & 233 $\pm$ 9 & 269 $\pm$ 9\\
48k & 212 $\pm$ 9 & 242 $\pm$ 9 & 279 $\pm$ 9 & 320 $\pm$ 10\\
56k & 247 $\pm$ 9 & 283 $\pm$ 9 & 326 $\pm$ 9 & 374 $\pm$ 9\\
64k & 283 $\pm$ 9 & 324 $\pm$ 9 & 372 $\pm$ 8 & 429 $\pm$ 9\\
128k & 551 $\pm$ 4 & 639 $\pm$ 7 & 725 $\pm$ 9 & 843 $\pm$ 8\\
\hline
\end{tabular}
}
\caption{Mean and standard deviation statistics of the numbers of tables in input prompts for our augmented long-context test sets with respect to each model's tokenizer.}
\label{tab:table_statistics}
\end{table}

\paragraph*{\textbf{Baseline Models and Evaluation Metrics}} 
We evaluate SQLong using two powerful base models: CodeQwen1.5-7B-Chat \citep{qwen}, which supports a context length of up to 64k, and Llama-3.1-8B-Instruct \citep{dubey2024llama}, which supports a context length of up to 128k. Following \citet{yu2018spider}, we report execution-match accuracy on both the original short-context test sets and the augmented long-context test sets.

\begin{table*}[!ht]
\centering
\scriptsize
\resizebox{0.99\textwidth}{!}{%
\def\arraystretch{1.2}
\begin{tabular}{l|cccccc}
\hline
\bf Model & \bf Spider-dev & \bf Spider-realistic & \bf Spider-syn & \bf Spider-test & \bf BIRD-dev & \bf Average \\
\hline
Qwen2-72B-Instruct & 82.7 & 80.7 & 73.0 & 82.9 & 53.7 & 74.6 \\
\hline
CodeQwen1.5-7B-Chat & 76.4 & 70.1 & 62.7 & 75.1 & 44.3 & 65.7 \\
\hdashline
\quad Finetuned without SQLong & 81.9 & 76.2 & 68.7 & 79.6 & 51.4 & 71.6 \\
\quad Finetuned with SQLong & \textbf{83.4} & \textbf{79.7} & \textbf{71.2} & \textbf{81.3} & \textbf{53.3} & \textbf{73.8} \\
\hline
\hline
Llama-3.1-70B-Instruct & 80.7 & 78.0 & 73.0 & 83.7 & 61.5 & 75.4 \\
\hline
Llama-3.1-8B-Instruct & 71.1 & 63.8 & 61.0 & 65.7 & 40.9 & 60.5 \\
\hdashline
\quad Finetuned without SQLong & 79.2 & 76.4 & 69.6 & 80.4 & 51.9 & 71.5 \\
\quad Finetuned with SQLong & \textbf{83.2} & \textbf{78.0} & \textbf{73.1} & \textbf{81.8} & \textbf{53.3} & \textbf{73.9} \\
\hline
\end{tabular}%
}
\caption{Execution-match accuracy results (in \%) across different datasets and model configurations. Finetuning with SQLong consistently improves performance, with the best results highlighted in \textbf{bold}.}
\label{tab:results_on_original_data}
\end{table*}

\paragraph*{\textbf{Training Protocol}} For each original training set, we use SQLong to create an augmented \textit{long-context finetuning} dataset with context lengths of up to 32k.\footnote{Due to computational constraints, we limit finetuning to context lengths of up to 32k. Specifically, for each training example, the context length is randomly sampled from a range starting at 4,096 and increasing by 512 increments up to 32,768.} The augmented dataset is combined with the original training set to form the final dataset used for finetuning the base models.\footnote{For Spider, we finetune the base models on the Spider training set and evaluate performance on Spider-dev, Spider-test, Spider-realistic, and Spider-syn.}

We experiment with two base models: CodeQwen1.5-7B-Chat \citep{qwen}, which supports a 64k context length, and Llama-3.1-8B-Instruct \citep{dubey2024llama}, which supports a 128k context length. Finetuning is performed with a batch size of 1, gradient accumulation steps of 8, a learning rate chosen from ${1 \times 10^{-6}, 5 \times 10^{-6}, 1 \times 10^{-5}}$, and up to 5 epochs on 8$\times$H100 80GB GPUs.

We use Huggingface's TRL \citep{vonwerra2022trl} for supervised finetuning, employing 8-bit AdamW \citep{dettmers20218}, Flash Attention v2 \citep{dao2023flashattention}, and DeepSpeed ZeRO-3 Offload \citep{ren2021zero}. For a fair comparison, we also finetune the base models on the original training set (i.e., without SQLong) under the same settings.

\paragraph*{\textbf{Inference Protocol}} We utilize vLLM \citep{kwon2023efficient} for the inference process. For long-context test sets, we employ dynamic NTK RoPE scaling \citep{peng2023yarn} to extend support up to a 128k context length for CodeQwen1.5-7B-Chat and its finetuned variants.

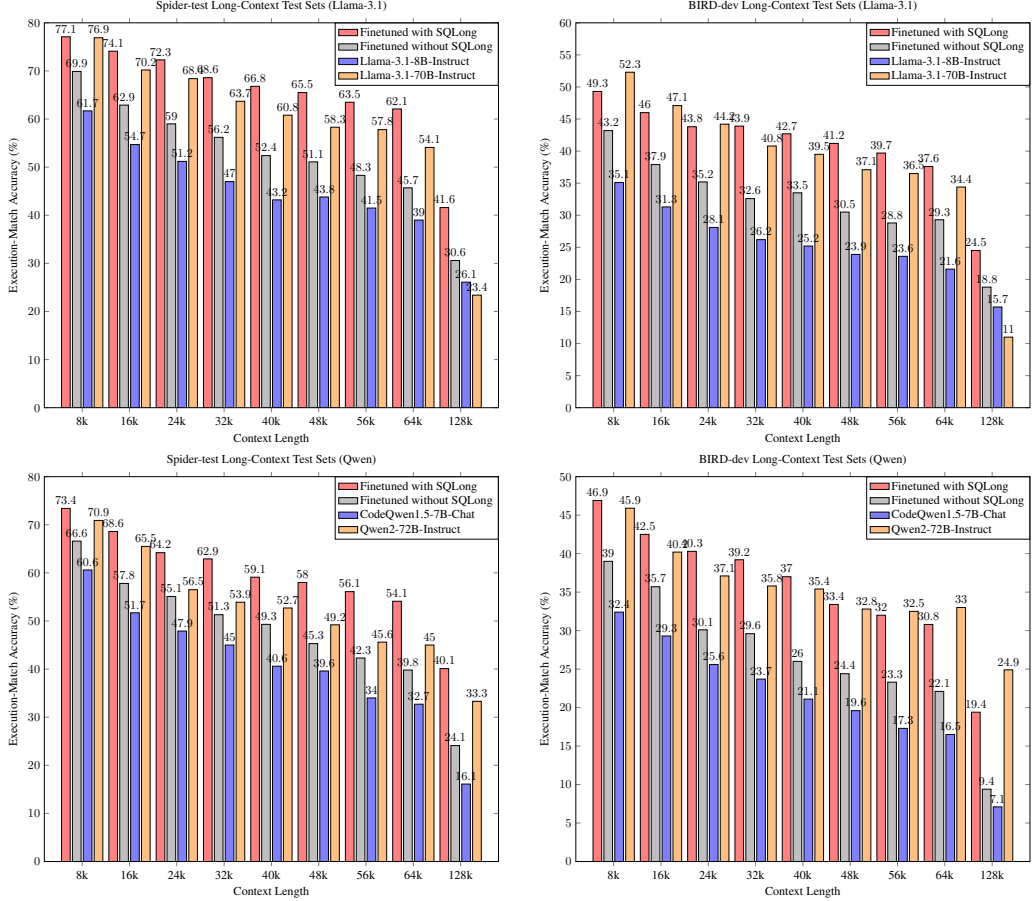
\begin{figure*}[!htb]
\centering
\begin{subfigure}[t]{\textwidth}
\centering
\resizebox{0.85\textwidth}{!}{%
\begin{tikzpicture}
\begin{axis}[
    title={Spider-test Long-Context Test Sets (Llama-3.1)},
    ybar,
    bar width=0.275cm,
    width=1\textwidth,
    enlarge x limits=0.1,
    legend style={at={(1,1)}, anchor=north east, legend columns=1, legend cell align=left},
    ylabel={Execution-Match Accuracy (\%)},
    xlabel={Context Length},
    symbolic x coords={8k, 16k, 24k, 32k, 40k, 48k, 56k, 64k, 128k},
    xtick=data,
    ymin=0, ymax=80,
    nodes near coords,
    area legend,
    ]

    \addplot[fill=red!50] coordinates {
        (8k, 77.1) (16k, 74.1) (24k, 72.3) (32k, 68.6) 
        (40k, 66.8) (48k, 65.5) (56k, 63.5) (64k, 62.1) (128k, 41.6)};
    \addplot[fill=gray!50] coordinates {
        (8k, 69.9) (16k, 62.9) (24k, 59.0) (32k, 56.2) 
        (40k, 52.4) (48k, 51.1) (56k, 48.3) (64k, 45.7) (128k, 30.6)};
    \addplot[fill=blue!50] coordinates {
        (8k, 61.7) (16k, 54.7) (24k, 51.2) (32k, 47.0) 
        (40k, 43.2) (48k, 43.8) (56k, 41.5) (64k, 39.0) (128k, 26.1)};
    \addplot[fill=orange!50] coordinates {
        (8k, 76.9) (16k, 70.2) (24k, 68.4) (32k, 63.7) 
        (40k, 60.8) (48k, 58.3) (56k, 57.8) (64k, 54.1) (128k, 23.4)};
    \legend{Finetuned with SQLong, Finetuned without SQLong, Llama-3.1-8B-Instruct, Llama-3.1-70B-Instruct}
\end{axis}
\end{tikzpicture}
\hspace{1cm}
\begin{tikzpicture}
\begin{axis}[
    title={BIRD-dev Long-Context Test Sets (Llama-3.1)},
    ybar,
    bar width=0.275cm,
    width=1\textwidth,
    enlarge x limits=0.1,
    legend style={at={(1,1)}, anchor=north east, legend columns=1, legend cell align=left},
    ylabel={Execution-Match Accuracy (\%)},
    xlabel={Context Length},
    symbolic x coords={8k, 16k, 24k, 32k, 40k, 48k, 56k, 64k, 128k},
    xtick=data,
    ymin=0, ymax=60,
    nodes near coords,
    area legend,
    ]

    \addplot[fill=red!50] coordinates {
        (8k, 49.3) (16k, 46.0) (24k, 43.8) (32k, 43.9) 
        (40k, 42.7) (48k, 41.2) (56k, 39.7) (64k, 37.6) (128k, 24.5)};
    \addplot[fill=gray!50] coordinates {
        (8k, 43.2) (16k, 37.9) (24k, 35.2) (32k, 32.6) 
        (40k, 33.5) (48k, 30.5) (56k, 28.8) (64k, 29.3) (128k, 18.8)};
    \addplot[fill=blue!50] coordinates {
        (8k, 35.1) (16k, 31.3) (24k, 28.1) (32k, 26.2) 
        (40k, 25.2) (48k, 23.9) (56k, 23.6) (64k, 21.6) (128k, 15.7)};
    \addplot[fill=orange!50] coordinates {
        (8k, 52.3) (16k, 47.1) (24k, 44.2) (32k, 40.8) 
        (40k, 39.5) (48k, 37.1) (56k, 36.5) (64k, 34.4) (128k, 11.0)};
    \legend{Finetuned with SQLong, Finetuned without SQLong, Llama-3.1-8B-Instruct, Llama-3.1-70B-Instruct}
\end{axis}
\end{tikzpicture}
}
\end{subfigure}

\begin{subfigure}[t]{\textwidth}
\centering
\resizebox{0.85\textwidth}{!}{%
\begin{tikzpicture}
\begin{axis}[
    title={Spider-test Long-Context Test Sets (Qwen)},
    ybar,
    bar width=0.275cm,
    width=1\textwidth,
    enlarge x limits=0.1,
    legend style={at={(1,1)}, anchor=north east, legend columns=1, legend cell align=left},
    ylabel={Execution-Match Accuracy (\%)},
    xlabel={Context Length},
    symbolic x coords={8k, 16k, 24k, 32k, 40k, 48k, 56k, 64k, 128k},
    xtick=data,
    ymin=0, ymax=80,
    nodes near coords,
    area legend,
    ]

    \addplot[fill=red!50] coordinates {
        (8k, 73.4) (16k, 68.6) (24k, 64.2) (32k, 62.9) 
        (40k, 59.1) (48k, 58.0) (56k, 56.1) (64k, 54.1) (128k, 40.1)};
    \addplot[fill=gray!50] coordinates {
        (8k, 66.6) (16k, 57.8) (24k, 55.1) (32k, 51.3) 
        (40k, 49.3) (48k, 45.3) (56k, 42.3) (64k, 39.8) (128k, 24.1)};
    \addplot[fill=blue!50] coordinates {
        (8k, 60.6) (16k, 51.7) (24k, 47.9) (32k, 45.0) 
        (40k, 40.6) (48k, 39.6) (56k, 34.0) (64k, 32.7) (128k, 16.1)};
    \addplot[fill=orange!50] coordinates {
        (8k, 70.9) (16k, 65.5) (24k, 56.5) (32k, 53.9) 
        (40k, 52.7) (48k, 49.2) (56k, 45.6) (64k, 45.0) (128k, 33.3)};
    \legend{Finetuned with SQLong, Finetuned without SQLong, CodeQwen1.5-7B-Chat, Qwen2-72B-Instruct}
\end{axis}
\end{tikzpicture}
\hspace{1cm}
\begin{tikzpicture}
\begin{axis}[
    title={BIRD-dev Long-Context Test Sets (Qwen)},
    ybar,
    bar width=0.275cm,
    width=1\textwidth,
    enlarge x limits=0.1,
    legend style={at={(1,1)}, anchor=north east, legend columns=1, legend cell align=left},
    ylabel={Execution-Match Accuracy (\%)},
    xlabel={Context Length},
    symbolic x coords={8k, 16k, 24k, 32k, 40k, 48k, 56k, 64k, 128k},
    xtick=data,
    ymin=0, ymax=50,
    nodes near coords,
    area legend,
    ]

    \addplot[fill=red!50] coordinates {
        (8k, 46.9) (16k, 42.5) (24k, 40.3) (32k, 39.2) 
        (40k, 37.0) (48k, 33.4) (56k, 32.0) (64k, 30.8) (128k, 19.4)};
    \addplot[fill=gray!50] coordinates {
        (8k, 39.0) (16k, 35.7) (24k, 30.1) (32k, 29.6) 
        (40k, 26.0) (48k, 24.4) (56k, 23.3) (64k, 22.1) (128k, 9.4)};
    \addplot[fill=blue!50] coordinates {
        (8k, 32.4) (16k, 29.3) (24k, 25.6) (32k, 23.7) 
        (40k, 21.1) (48k, 19.6) (56k, 17.3) (64k, 16.5) (128k, 7.1)};
    \addplot[fill=orange!50] coordinates {
        (8k, 45.9) (16k, 40.2) (24k, 37.1) (32k, 35.8) 
        (40k, 35.4) (48k, 32.8) (56k, 32.5) (64k, 33.0) (128k, 24.9)};
    \legend{Finetuned with SQLong, Finetuned without SQLong, CodeQwen1.5-7B-Chat, Qwen2-72B-Instruct}
\end{axis}
\end{tikzpicture}
}
\end{subfigure}
\caption{Execution-match accuracy (in \%) for Llama-3.1 (top) and Qwen (bottom) families on Spider-test (left) and BIRD-dev (right) long-context test sets.}
\label{fig:combined_results}
\end{figure*}

\subsection{Main Results}

\paragraph*{\textbf{Performance on Original Datasets}} 
Table \ref{tab:results_on_original_data} summarizes the results on the original development and test sets, comparing base models with larger LLMs such as Llama-3.1-70B-Instruct \citep{dubey2024llama} and Qwen2-72B-Instruct \citep{qwen2}. Models finetuned using long-context augmentation via SQLong consistently outperform their counterparts finetuned on original contexts. On average, SQLong delivers an absolute improvement of over 2.2\% across five benchmark datasets. Additionally, SQLong-finetuned models achieve performance comparable to much larger LLMs on specific datasets, showcasing the scalability and efficiency of the approach.

\paragraph*{\textbf{Performance on Long-Context Datasets}} 
Figure~\ref{fig:combined_results} illustrates the experimental results on long-context test sets. 
The full details are presented in Tables \ref{tab:results_on_original_data_llama} and \ref{tab:results_on_original_data_codeqwen} in Appendix \ref{subsec:full_long_context_results}.
Across all datasets, models finetuned with SQLong demonstrate superior performance compared to those trained without SQLong. 
For instance, on the Spider-test datasets with 8k and 24k context lengths, the Llama-3.1-8B-Instruct model achieves outstanding results of 77.1\% and 72.3\%, reflecting absolute gains of 7.2\% and 13.3\%, respectively. Notably, the SQLong-finetuned Llama-8B model outperforms the larger Llama-70B model on 41 out of 45 long-context test sets, with minor exceptions on Spider-realistic 8k and BIRD-dev 8k, 16k, and 24k sets. Similar performance trends are observed with the Qwen models.

On average, SQLong finetuning delivers an 11\% absolute improvement over models without SQLong and a 6\% advantage over 70B models within the same model family. These results underscore the efficacy of SQLong in handling long-context scenarios and advancing the performance of NL2SQL systems. 

\begin{figure}[!ht]
\centering
\includegraphics[width=0.45\textwidth]{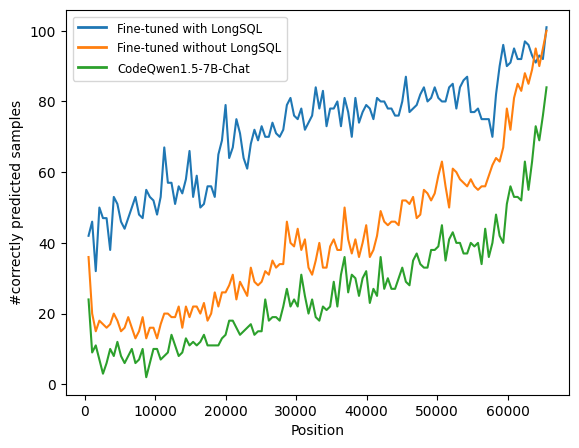}
\caption{Robust impact of fine-tuned models.}
\label{fig:robust_impact}
\end{figure}

\paragraph*{\textbf{Positional robustness}} We conduct an experiment wherein each original database schema is placed at different positions within the input prompt, assessing the models' ability to detect it regardless of its location.

We select a set of 124 samples from Spider-dev, Spider-realistic, and Spider-syn, ensuring each sample has a maximum input prompt and target SQL query length of 384 tokens according to CodeQwen1.5-7B-Chat's tokenizer. Using SQLong, we augment this set to a 64k context length. In each augmented set, the original database schemas are positioned at specific offsets, starting from 512 and incrementing by 512 up to 64k. This results in 125 new test sets, each containing 124 samples with a 64k context length, corresponding to a distinct schema position.

We compute the number of correctly executed samples for each test set, as shown in Figure \ref{fig:robust_impact}. The results demonstrate that the long-context fine-tuned model with SQLong is significantly more robust compared to the model without fine-tuning.

\section{Conclusion and Future Work}
Handling large database schemas poses a significant challenge for NL2SQL models. In this paper, we introduce long-context NL2SQL generation, a novel task that reflects real-world scenarios, and propose SQLong, a simple yet effective augmentation approach for creating long-context finetuning and benchmark datasets. Experiments show that LLMs finetuned with SQLong significantly outperform their counterparts on benchmarks like Spider, BIRD, and our long-context test sets (up to 128k context length). 

Future work includes leveraging a RAG-based schema linking approach to retrieve relevant schema elements, enabling more concise and efficient inputs for SQLong-tuned models.

\bibliography{references}

\appendix

\section{Appendix}
\label{sec:appendix}

\subsection{The algorithm steps in SQLong}
\label{subsec:SQLong_algorithm}

\begin{algorithm*}[!ht]
\caption{The algorithm steps involved in the proposed SQLong.}
\label{alg:SQLong_algorithm}
\DontPrintSemicolon
\SetAlgoVlined

\textbf{Input}: A training set $\mathbf{T}$ of pairs of input prompts and target SQL queries: $\mathbf{T} = \{((instructions_i, database\_schema_i, question_i), target\_sql_i)\}_{i=1}^N$, wherein each $database\_schema_i$ is a set of $\mathsf{CREATE\ TABLE}$ commands and three data rows for each corresponding table; a set $\mathcal{T} = \{((instructions_j, database\_schema_j, question_j), target\_sql_j)\}_{j=1}^M$; the base model's tokenizer $tk$, a starting number $s\_n$ (default 4096), an ending number $e\_n$ (default 32768), an increasing number $i\_n$ (default 512), and a pre-defined number $p\_n$ (default 8192).

\textbf{Output}: The augmented long-context set $\mathcal{T}'$.

$schema\_set \leftarrow$ collect\_unique\_commands\_and\_data\_rows($\{database\_schema_i\}_{i=1}^N$)

$table\_names \leftarrow$ get\_table\_names($schema\_set$)

$item\_lengths \leftarrow \{\}$

\For{$item \in schema\_set$}{

	$item\_lengths \leftarrow item\_lengths\ \cup$ \{get\_length($item$, $tk$)\}

}

$\mathcal{T}' \leftarrow \{\}$

$diverse\_lengths \leftarrow$ range($s\_n$, $e\_n+1$, $i\_n$) 

\For{$((instructions, database\_schema, question), target\_sql) \in \mathcal{T}$}{
	$original\_length \leftarrow$ get\_length($instructions + database\_schema + question + target\_sql$, $tk$)
	
	$certain\_length \leftarrow$ randomly\_select\_value($diverse\_lengths$) \tcp*{This aims to construct long-context fine-tuning data with $\mathbf{T} = \mathcal{T}$. Otherwise, $certain\_length$ is set to $p\_n$ to construct long-context benchmark data.}
	
	$local\_table\_names \leftarrow$ get\_table\_names($database\_schema$)
	
	$augmented\_schema \leftarrow \{\}$
	
	\For{$idx \in \mathsf{shuffle\_list(range(}0, \mathsf{get\_size(}schema\_set\mathsf{)))}$}{
	
		\If{$schema\_set[idx] \notin database\_schema\ \mathsf{and}\ table\_names[idx] \notin local\_table\_names\ \mathsf{and}\ original\_length + item\_lengths[idx] < certain\_length$} { 
		
			$original\_length \leftarrow original\_length\ + item\_lengths[idx]$
			
			$augmented\_schema \leftarrow augmented\_schema\ \cup \{schema\_set[idx]\}$
			
		}
	
	}
	
	$augmented\_long\_context\_schema \leftarrow$ shuffle\_list($augmented\_schema\ \cup database\_schema$)
	
	$\mathcal{T}' \leftarrow \mathcal{T}' \cup \{((instructions, augmented\_long\_context\_schema, question), target\_sql)\}$
}
\end{algorithm*}

\newpage

\subsection{Full execution-match accuracy results for all long-context test sets}
\label{subsec:full_long_context_results}

\begin{table*}[!ht]
\centering
\resizebox{16cm}{!}{
\def\arraystretch{1.05}
\begin{tabular}{l|c|ccccc|c}
\hline
\bf Model &\bf Context &  & & \bf Dataset & & & \bf Average\\
& \bf length & \bf Spider-dev & \bf Spider-realistic & \bf Spider-syn & \bf Spider-test & \bf BIRD-dev & \bf across 45 sets\\
\hline
Llama-3.1-8B-Instruct & 8k &	61.9 & 	 53.5 & 	45.1 & 	61.7 &	35.1 \\
& 16k &	58.5 & 	47.0 & 	38.9 & 	54.7 &	31.3 \\
& 24k &	53.2 & 	 43.1 & 	32.7 & 	51.2 &	28.1 \\
& 32k &	49.6 & 	42.9 & 	29.9 & 	47.0 &	26.2 \\
& 40k &	48.7 & 	 38.4 & 	28.4 & 	43.2 &	25.2 & 37.2\\
& 48k &	46.9 & 	 35.8 & 	24.9 &	43.8 &	23.9 \\
& 56k &	45.5 & 	 32.1 & 	23.8 & 	41.5 &	23.6 \\
& 64k &	42.6 & 	 33.1 & 	22.5 & 	39.0 &	21.6 \\
& 128k &	28.0 &	17.9 &	10.3 &	26.1 &	15.7 \\
\hline
Our model fine-tuned & 8k &	71.7 & 	 63.4 & 	49.3 & 	69.9 &	43.2 \\
\ \ \ \ \ Without SQLong& 16k &	66.6 & 	 54.7 & 	39.9 & 	62.9 &	37.9 \\
& 24k &	63.6 &	 52.4 & 	35.5 & 	59.0 &	35.2 \\
& 32k &	59.4 & 	 48.0 & 	33.1 & 	56.2 &	32.6 \\
& 40k &	57.0 & 	 45.1 & 	30.2 & 	52.4 &	33.5 & 43.8\\
& 48k &	55.9 & 	 43.7 & 	28.0 & 	51.1 &	30.5 \\
& 56k &	52.5 & 	 40.4 & 	25.7 & 	48.3 &	28.8 \\
& 64k &	51.4 & 	 40.9 & 	25.3 & 	45.7 &	29.3 \\
& 128k &	34.7 &	23.6 &	13.5 &	30.6 &	18.8\\
\hline
Our model fine-tuned &  8k & \bf	77.4 & 	 67.1 & \bf 	 61.7 & \bf	77.1 &	49.3 \\
\ \ \ \ \ With SQLong& 16k & \bf 	75.2 & \bf 	 66.1 & \bf 	 53.4 & \bf 	 74.1 &	46.0 \\
& 24k & \bf	71.8 & \bf 	 64.2 & \bf 	 50.0 & \bf 	 72.3 &	43.8 \\
& 32k & \bf	68.3 & \bf 	 61.6 & \bf 	 46.5 & \bf 	 68.6 & \bf	43.9 \\
& 40k & \bf	67.5 & \bf 	 62.8 & \bf 	 44.9 & \bf 	 66.8 & \bf	42.7 & \bf 54.8\\
& 48k & \bf	66.9 & \bf 	 56.7 & \bf 	 40.2 & \bf 	 65.5 & \bf	41.2 \\
& 56k & \bf	63.3 & \bf	 52.6 & \bf 	 38.4 & \bf	 63.5 & \bf	39.7 \\
& 64k & \bf	61.3 & \bf 	 52.2 & \bf 	 39.3 & \bf 	 62.1 & \bf	37.6 \\
& 128k & \bf	43.0 & \bf	33.7 & \bf	21.7 & \bf	41.6 & \bf	24.5\\
\hline
\hline
Llama-3.1-70B-Instruct &  8k &	73.9 & \bf 	67.3 & 	55.0 & 	76.9 & \bf	52.3 \\
& 16k &	67.7 & 	59.4 & 	48.9 & 	70.2 & \bf	47.1 \\
& 24k &	62.4 & 	54.9 & 	43.8 & 	68.4 & \bf	44.2 \\
& 32k &	60.9 & 	49.6 & 	41.7 & 	63.7 &	40.8 \\
& 40k &	59.0 & 	52.6 & 	37.4 & 	60.8 &	39.5 & 48.5\\
& 48k &	57.6 & 	46.9 & 	35.0 & 	58.3 &	37.1 \\
& 56k &	55.3 & 	46.3 & 	32.3 & 	57.8 &	36.5 \\
& 64k &	55.0 & 	43.9 &	31.7 & 	54.1 &	34.4 \\
& 128k &	28.0 &	25.6 &	12.3 &	23.4 &	11.0\\
\hline
\end{tabular}
}
\caption{Execution-match accuracy results (in \%) on the augmented long-context test sets with respect to the Llama-3.1 model family.}
\label{tab:results_on_original_data_llama}
\end{table*}

\begin{table*}[!ht]
\centering
\resizebox{16cm}{!}{
\def\arraystretch{1.05}
\begin{tabular}{l|c|ccccc|c}
\hline
\bf Model &\bf Context &  & & \bf Dataset & & & \bf Average\\
& \bf length & \bf Spider-dev & \bf Spider-realistic & \bf Spider-syn & \bf Spider-test & \bf BIRD-dev & \bf across 45 sets\\
\hline
CodeQwen1.5-7B-Chat & 8k &	61.7 & 	49.6 & 	38.1 & 	60.6 &	32.4 \\
& 16k &	55.9 & 	42.1 & 	30.7 &	51.7 &	29.3 \\
& 24k &	51.5 & 	37.8 & 	27.9 & 	47.9 &	25.6 \\
& 32k &	48.0 & 	30.9 & 	22.8 & 	45.0 &	23.7 \\
& 40k &	46.7 & 	28.9 & 	21.0 & 	40.6 &	21.1 & 31.7\\
& 48k &	42.4 & 	27.8 & 	18.7 & 	39.6 &	19.6 \\
& 56k &	36.4 & 	24.0 & 	17.5 & 	34.0 &	17.3 \\
& 64k &	36.4 & 	21.3 & 	15.8 & 	32.7 &	16.5 \\
& 128k &	19.2 & 	7.9 & 	6.4 & 	16.1 &	7.1\\
\hline
Our model fine-tuned & 8k &	68.9 & 	57.1 & 	39.5 & 	66.6 &	39.0 \\
\ \ \ \ \ Without SQLong & 16k &	62.6 & 	51.4 & 	31.8 & 	57.8 &	35.7 \\
& 24k &	57.6 & 	 49.0 & 	29.3 & 	55.1 &	30.1 \\
& 32k &	53.0 & 	41.5 &  	25.6 & 	51.3 &	29.6 \\
& 40k &	53.7 & 	38.4 & 	23.5 & 	49.3 &	26.0 & 37.8\\
& 48k &	48.7 & 	34.6 & 	22.3 & 	45.3 &	24.4 \\
& 56k &	44.5 & 	33.1 & 	20.9 & 	42.3 &	23.3 \\
& 64k &	43.8 & 	30.3 & 	18.4 & 	39.8 &	22.1 \\
& 128k &	26.1 & 	15.6 & 	9.2 &	24.1 &	9.4 \\
\hline
Our model fine-tuned & 8k & \bf	75.9 & \bf 	65.7 & \bf 	53.2 & \bf 	73.4 & \bf	46.9 \\
\ \ \ \ \ With SQLong & 16k & \bf	72.9 & \bf 	62.6 & \bf 	46.6 & \bf 	68.6 & \bf	42.5 \\
& 24k & \bf	68.9 & \bf 	58.5 & \bf 	43.0 & \bf 	64.2 & \bf	40.3 \\
& 32k & \bf	67.5 & \bf 	54.3 & \bf 	40.0 & \bf 	62.9 & \bf	39.2 \\
& 40k & \bf	63.4 & \bf 	53.7 & \bf 	37.4 & \bf 	59.1 & \bf	37.0 & \bf 50.2\\
& 48k & \bf	63.9 & \bf 	52.8 & \bf 	35.3 & \bf 	58.0 & \bf	33.4 \\
& 56k & \bf	60.3 & \bf	51.0 & \bf 	33.6 & \bf 	56.1 &	32.0 \\
& 64k & \bf	60.6 & \bf 	52.4 & \bf 	31.0 & \bf 	54.1 &	30.8 \\
& 128k & \bf	43.4 & \bf 	33.7 & \bf 	19.4 &  \bf	40.1 &	19.4 \\
\hline
\hline
Qwen2-72B-Instruct &  8k &	70.6 & 	63.4 & 	47.2 & 	70.9 &	45.9 \\
& 16k &	69.1 & 	58.7 & 	40.6 & 	65.5 &	40.2 \\
& 24k &	60.9 & 	53.3 & 	34.1 & 	56.5 &	37.1 \\
& 32k &	59.6 & 	45.5 & 	31.1 & 	53.9 &	35.8 \\
& 40k &	55.8 & 	45.7 & 	29.5 & 	52.7 &	35.4 & 44.2\\
& 48k &	52.3 &	43.7 & 	27.8 & 	49.2 &	32.8 \\
& 56k &	50.8 & 	39.4 & 	27.6 & 	45.6 & \bf	32.5 \\
& 64k &	47.3 & 	34.6 & 	25.1 & 	45.0 & \bf	33.0 \\
& 128k & 36.8 & 28.3 & 18.6 & 33.3 & \bf 24.9	\\
\hline
\end{tabular}
}
\caption{Execution-match accuracy results (in \%) on the augmented long-context test sets with respect to the Qwen mdoel family.}
\label{tab:results_on_original_data_codeqwen}
\end{table*}

\end{document}